\documentclass[11pt,a4paper]{article}
\usepackage[hyperref]{acl2019}
\usepackage{booktabs}
\usepackage{times}
\usepackage{latexsym}
\usepackage{comment}
\usepackage{xcolor}
\usepackage{graphicx}
\usepackage{stfloats}
\usepackage{multicol}
\usepackage{amsmath,amsfonts,amssymb}
\usepackage{multirow}
\usepackage{url}
\usepackage{bm}

\aclfinalcopy % Uncomment this line for the final submission
\def\aclpaperid{37} %  Enter the acl Paper ID here

\title{Probing Multilingual Sentence Representations With X-P\textsc{robe}}

\author{Vinit Ravishankar \\
  Language Technology Group \\
  Department of Informatics \\
  University of Oslo \\
  \texttt{vinitr@ifi.uio.no} \\\And
  Lilja {\O}vrelid \\
  Language Technology Group \\
  Department of Informatics \\
  University of Oslo \\
  \texttt{liljao@ifi.uio.no} \\\And
  Erik Velldal \\
  Language Technology Group \\
  Department of Informatics \\
  University of Oslo \\
  \texttt{erikve@ifi.uio.no} \\}

\date{}

\begin{document}
\maketitle
\begin{abstract}
This paper extends the task of probing sentence representations for linguistic insight in a multilingual domain. In doing so, we make two contributions: first, we provide datasets for multilingual probing, derived from Wikipedia, in five languages, viz. English, French, German, Spanish and Russian. Second, we evaluate six sentence encoders for each language, each trained by mapping sentence representations to English sentence representations, using sentences in a parallel corpus. We discover that cross-lingually mapped representations are often better at retaining certain linguistic information than representations derived from English encoders trained on natural language inference (NLI) as a downstream task.
\end{abstract}

\section{Introduction}

In recent years, there has been a considerable amount of research into attempting to represent contexts longer than single words with fixed-length vectors. These representations typically tend to focus on attempting to represent sentences, although phrase- and paragraph-centric mechanisms do exist. These have moved well beyond relatively na\"{i}ve compositional methods, such as additive and multiplicative methods \citep{mitchell_vector-based_2008}, one of the earlier papers on the subject. There have been several proposed approaches to learning these representations since, both unsupervised and supervised. Naturally, this has also sparked interest in evaluation methods for sentence representations; the focus of this paper is on \textit{probing}-centric evaluations, and their extension to a multilingual domain.

In Section~\ref{sec:bg}, we provide a literature review of prior work in the numerous domains that our paper builds upon. Section~\ref{sec:mle} motivates the principle of cross-lingual probing and describes our goals. In Section~\ref{sec:probe}, we describe our probing tasks and relevant modifications, if any. Section~\ref{sec:encoders} describes our sentence encoders, as well as the procedure we follow for training, mapping and probing. Section~\ref{sec:data} describes our data and relevant preprocessing methods we applied. Section~\ref{sec:eval} presents a detailed evaluation from several perspectives, which we discuss in Section~\ref{sec:disc}. We conclude, as well as describe avenues for future work, in Section~\ref{sec:conc}. Our hyperparameters are described in Appendix~\ref{sec:hyper}, and further detailed results that are not critical to the paper are tabulated in~\ref{sec:add}.

\section{Background}
\label{sec:bg}

\subsection{Sentence representation learning}
Numerous methods for learning sentence representations exist. Many of these methods are unsupervised, and thus do not have much significant annotation burden. Most of these methods are, however, structured: they rely on the sentences in training data being ordered and not randomly sampled. The aptly named \textit{SkipThoughts} \citep{kiros_skip-thought_2015} is a well-known earlier work, and uses recurrent encoder-decoder models to `decode' sentences surrounding the encoded sentence, using the final encoder state as the encoded sentence's representation. \citet{cer_universal_2018} evaluate two different encoders, a deep averaging network and a transformer, on unsupervised data drawn from a variety of web sources. \citet{hill_learning_2016} describe a model based on denoising auto-encoders, and a simplified variant of SkipThoughts, that sums up source word embeddings, that they dub (\textit{FastSent}). Another SkipThoughts variant \citep{logeswaran_efficient_2018} uses a multiple-choice objective for contextual sentences, over the more complicated decoder-based objective.

Several supervised approaches to building representations also exist. An earlier work is Charagram~\citep{wieting_charagram:_2016}, which uses paraphrase data and builds on character representations to arrive at sentence representations. More recent papers use a diverse variety of target tasks to ground representations, such as visual data~\citep{kiela_learning_2017}, machine translation data~\citep{mccann_learned_2017}, and even multiple tasks, in a multi-task learning framework \citep{subramanian_learning_2018}. Relevant to this paper is \citeauthor{conneau_supervised_2017}'s (\citeyear{conneau_supervised_2017}) \textit{InferSent}, that uses natural language inference (NLI) data to ground representations: they learn these representations on the well-known SNLI dataset~\citep{bowman_large_2015}.

\subsection{Multilingual representations}
Whilst sentence representation is a thriving research domain, there has been relatively less work on multilingualism in the context of sentence representation learning: most prior work has been focussed on multilingual word representation. For sentence representations, an early work \citep{schwenk_learning_2017} proposes a seq2seq-based objective, using machine learning encoders to map source sequences to fixed-length vectors. Along similar lines, \citet{conneau_xnli:_2018} propose using machine translation data to transfer sentence representations pre-trained on NLI, using a mean squared error (MSE) loss - this is the approach we follow. 

\citet{artetxe_massively_2018} present a `language agnostic' sentence representation system learnt over machine translation; the agnosticism refers to the joint BPE vocabulary that they construct over all languages, giving their encoders no language information, whilst their decoders are told what language to generate. Similarly,~\citet{lample_cross-lingual_2019} present pretrained cross-lingual models (XLM), based on modern pretraining mechanisms; specifically, a variant of the masked LM pretraining scheme used in BERT~\citep{devlin2018bert}.

Contemporaneous with this work, \citet{aldarmaki_scalable_2019} present an evaluation of three cross-lingual sentence transfer methods. Their methods include joint cross-lingual modelling methods that extend monolingual objectives to cross-lingual training, representation transfer learning methods that attempt to `optimise' sentence representations to be similar to parallel representations in another language, and sentence mapping methods based on orthogonal word embedding transfer: the authors use a parallel corpus as a `seed dictionary' to fit a transformation matrix between their source and target languages.

\subsection{On evaluation}

Work on evaluating sentence representations was encouraged by the release of the SentEval toolkit~\citep{conneau_senteval:_2018}, which provided an easy-to-use framework that sentence representations could be `plugged' into, for rapid downstream evaluation on numerous tasks: these include several classification tasks, textual entailment and similarity tasks, a paraphrase detection task, and caption/image retrieval tasks. \citet{conneau_what_2018} also created a set of `probing tasks', a variant on the theme of diagnostic classification~\citep{hupkes_visualisation_2017,belinkov_what_2017}, that would attempt to quantify precisely what sort of linguistic information was being retained by sentence representations. The authors, whose work focussed on evaluating representations for English, provided Spearman correlations between the performance of a particular representation mechanism on being probed for specific linguistic properties, and the downstream performance on a variety of NLP tasks. Along similar lines, and contemporaneously with this work,~\citet{liu_linguistic_nodate} probe three pretrained contextualised word representation models -- ELMo~\citep{peters2018deep}, BERT~\citep{devlin2018bert} and the OpenAI transformer~\citep{radford2018improving} -- with a ``suite of sixteen diverse probing tasks''.

On a different note,~\citet{saphra_understanding_2018} present a CCA-based method to compare representation learning dynamics across time and models, without explicitly requiring annotated probing corpora. They motivate the use of SVCCA~\citep{raghu2017svcca} to quantify precisely what an encoder learns by comparing the representations it generates with representations generated by an architecture trained specifically for a certain task, with the intuition that a higher similarity between the representations generated by the generic encoder and the specialised representations would indicate that the encoder is capable of encapsulating more task-relevant information. Their method has numerous advantages over traditional diagnostic classification, such as the elimination of the classifier, which reduces the risk of an additional component obfuscating results.

A visible limitation of the datasets provided by these probing tasks is that most of them were created with the idea of evaluating representations built for English language data. In this spirit, what we propose is analogous to \citeauthor{abdou_mgad:_2018}'s (\citeyear{abdou_mgad:_2018}) work on generating multilingual evaluation corpora for word representations. Within the realm of evaluating \textit{multilingual} sentence representations, \citet{conneau_xnli:_2018} describe the XNLI dataset, a set of translations of the development and test portions of the multi-genre MultiNLI inference dataset~\citep{williams_broad-coverage_2018}. This, in a sense, is an extension of a predominantly monolingual task to the multilingual domain; the authors evaluate sentence representations derived by \textit{mapping} non-English representations to an English representation space.

The original XNLI paper provides a baseline representation mapping technique, based on minimising the mean-squared error (MSE) loss between sentence representations across a parallel corpus. Their English language sentence representations are derived from an encoder trained on NLI data~\citep{conneau_supervised_2017}, and their target language representations are randomly initialised for a parallel sentence. While this system does perform reasonably well, a more naive machine-translation based approach performs better.

\section{Multilingual evaluation}
\label{sec:mle}
The focus of this paper is twofold. First, we provide five datasets for probing mapped sentence representations, in five languages (including an additional dataset for English), drawn from a different domain to \citeauthor{conneau_what_2018}'s probing dataset: specifically, from Wikipedia. Second, we probe a selection of mapped sentence representations, in an attempt to answer precisely what linguistic features are retained, and to what extent, post mapping. The emphasis of this evaluation is therefore, crucially, not a probing-oriented analysis of representations \textit{trained} on different languages, but an analysis of the effects of MSE-based mapping procedures on the ability of sentence representations to retain linguistic features. In this sense, our focus is less on the correlation between probing performance and downstream performance, and more on the relative performance of our representations on probing tasks.

Despite having described (in Section~\ref{sec:bg}) numerous methods, both for learning monolingual sentence representations, and for mapping them cross-linguistically, we restrict our work to a smaller subset of these. Specifically, we evaluate six encoders, each trained in a supervised fashion on NLI data.

Whilst our choice of languages could have been more typologically diverse, we were restricted by three factors:
\begin{enumerate}
    \item the availability of a parallel corpus with English for our mapping procedure
    \item the availability of a large enough Wikipedia to allow us to extract sufficient data (for instance, the Arabic Wikipedia was not large enough to fully extract data for all our tasks)
    \item the inclusion of the language in XNLI. Despite not being necessary, we believed it would be interesting to have a `real' downstream task to compare to.
\end{enumerate}

\section{Probing}
\label{sec:probe}
We use most of the probing tasks described in \citet{conneau_what_2018}. Due to the differences in corpus domain, we alter some of their word-frequency parameters. We also exclude the top constituent (\textbf{TopConst}) task; we noticed that Wikipedia tended to have far less diversity in sentence structure than the original Toronto Books corpus, due to the more encyclopaedic style of writing. A brief description of the tasks follows, although we urge the reader to refer to the original paper for more detailed descriptions. 

\begin{enumerate}
\item {Sentence length}:
In \textbf{SentLen}, sentences are divided into multiple bins based on their length; the job of the classifier is to predict the appropriate bin, creating a 6-way classification task.

\item {Word count}:
In \textbf{WC}, we sample sentences that feature exactly one amongst a thousand mid-frequency words, and train the classifier to predict the word: this is the most `difficult' task, in that it has the most possible classes.

\item {Tree depth}:
The \textbf{TreeDepth} task simply asks the representation to predict the depth of the sentence's syntax tree. Unlike the original paper, we use the depth the of the dependency tree instead of the constituency tree: this has the added benefit of being faster to extract, due to the relative speed of dependency parsing, as well as having better multilingual support. We also replace the authors' sentence-length-decorrelation procedure with a na\"{i}ver one, where we sample an equal number of $d$-depth trees for each sentence length bin.

\item {Bigram shift}:
In \textbf{BiShift}, for half the sentences in the dataset, the order of words in a randomly sampled bigram is reversed. The classifier learns to predict whether or not the sentence contains a reversal. 

\item {Subject number}:
The \textbf{SubjNum} task asks the classifier to predict the number of the subject of the head verb of the sentence. Only sentences with exactly one subject (annotated with the \texttt{nsubj} relation) attached to the root verb were considered.

\item {Object number}:
\textbf{ObjNum}, similar to the subject number task, was annotated with the number of the direct object of the head verb (annotated with the \texttt{obj} relation).

\item {Coordination inversion}:
In \textbf{CoordInv}, two main clauses joined by a coordinating conjunction (annotated with the \texttt{cc} and \texttt{conj} relations) have their orders reversed, with a probability of one in two. Only sentences with exactly two top-level conjuncts are considered.

\item {(Semantic) odd man out}:
\textbf{SOMO}, one of the more difficult tasks in the collection, replaces a randomly sampled word with another word with comparable corpus bigram frequencies, for both bigrams formed with the preceding and the succeeding words. We defined `comparable' as having a log-frequency difference not greater than 1.

\item {Tense prediction}:
The \textbf{Tense} prediction asks the classifier to predict the tense of the main verb: the task uses a rather simple division of tenses; two tenses, \texttt{Past} and \texttt{Pres}. Tense information was extracted from UD morphological annotation.
\end{enumerate}

\section{Encoders}
\label{sec:encoders}
\begin{figure*}[t]
    \centering
    \includegraphics[width=0.9\textwidth]{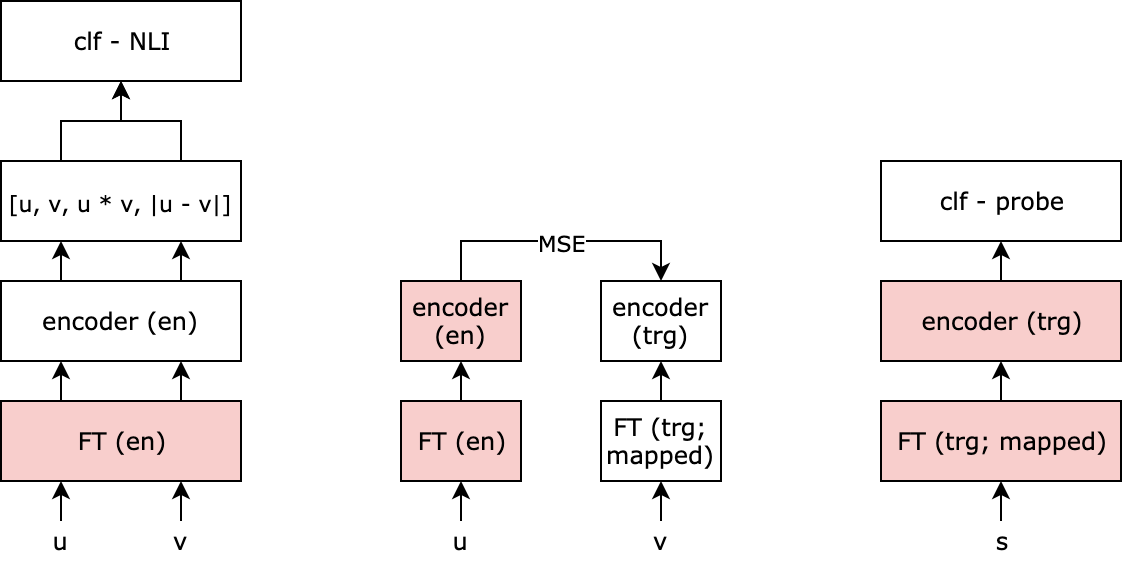}
    \caption{(a) an English-language encoder is trained on NLI data; (b) parallel sentences are encoded in English and the target language, and the MSE loss between them is minimised; (c) the mapped target encoders are used downstream in probing. Greyed-out blocks represent `frozen' components that do not further adjust their parameters.}
    \label{fig:nli}
\end{figure*}

The NLI-oriented training approach for all our encoders is based on \textit{InferSent}~\citep{conneau_supervised_2017}. Our first encoder is an RNN-based encoder (specifically, an LSTM); we use two variants of this encoder, one that uses max-pooling over bidirectional RNN states, and another that uses the last recurrent state. Our second encoder is a self-attention based encoder \citet{lin_structured_2017}, with the same max-pool/last-state variants. Finally, we include a convolutional sentence representation model inspired by~\citet{gan_learning_2016}; this model has an order of magnitude fewer parameters than the RNN- and attention-based variants. A variant of this CNN-based encoder has the maximum pooling replaced with average pooling.

\subsection{Representation learning}
\label{ssec:replearn}
We train all our encoders to represent sentences using the same NLI-based objective followed by~\citet{conneau_supervised_2017}. More precisely, we first convert the word indices for both our premise and our hypothesis into dense word representations using pretrained fastText word embeddings~\citep{bojanowski_enriching_2016}. These representations are then passed to our encoder of choice, which returns two fixed-length vectors: $u$ for the premise, and $v$ for the hypothesis. These vectors are combined and concatenated, as $[u, v, u * v, \mid u - v \mid]$, and then passed through a classifier with a softmax layer that outputs a probability distribution over the three NLI labels.

\subsection{Mapping}

Our procedure for mapping our encoders cross-linguistically broadly follows the principled mapping approach described in~\citet{conneau_xnli:_2018}. The procedure begins by mapping our \textit{word} representations to the same vector space. Unlike the original paper, we use the supervised variant of VecMap~\citep{artetxe_learning_2016} for representation mapping; however, we use seed dictionaries described in~\citet{conneau_word_2017}. Having mapped our word representations, we proceed to map our sentence representations. To do so, we first build an English-language encoder, using (frozen) word representations and (frozen) encoder weights obtained in Section~\ref{ssec:replearn}. We then build a target language encoder, using word embeddings mapped to the same space as the English embeddings. The sentence encoder itself is initialised with random parameters.

We then encode the source and target sentences in an en-trg machine translation corpus, where trg is our target language. Our English encoder returns a `meaningful' representation: recall that the encoder has weights trained on NLI data. We then use a mean-square error loss function to reduce the distance between our target-language representation and the English representation; the system then backpropagates through the target language encoder to obtain better parameters. 

Our MSE loss function, similar to \citeauthor{conneau_xnli:_2018}'s function, attempts to minimise the distance between representations of parallel sentences, whilst simultaneously maximising the distance between random sentences sampled from either language in the pair. Mathematically, the alignment loss is given by: $$\mathcal{L}_{align} = ||\mathbf{x} - \mathbf{y}||_2 - \lambda(||\mathbf{x_c} - \mathbf{y}||_2 + ||\mathbf{x} - \mathbf{y_c}||_2)$$

where $\lambda$ is a hyperparameter.

We evaluate our mapped encoder on the relevant validation data section from the XNLI corpus per epoch, and terminate the mapping procedure when our validation accuracy does not improve for two consecutive epochs. 

\subsection{Multilingual probing}
Having obtained our mapped sentence representation encoder, we proceed to plug the encoder into our probing architecture downstream, and evaluate classifier performance. 

First, we load our mapped word representations for the language that we intend to analyse. We use these word representations to build sentence representations, using the encoder architecture of choice. We then add a simple multi-layer perceptron (MLP) that learns to predict the appropriate label for each task: the MLP consists of a dense layer, a non-linearity (we use the sigmoid function), and another dense layer that we softmax over to arrive at per-class probabilities. During training, we keep the encoder's parameters fixed. Mathematically, therefore, given an encoder $f$ with parameters $\theta$, and word representations $\bm{w_k}$ for each word $k$:
\begin{align*}
   \bm{s} &= f(\bm{w_0}, \bm{w_1}, ..., \bm{w_n}; \theta)\\
   \bm{z} &= \mathrm{MLP}(s)\\
   \bm{y} &= \mathrm{softmax}(\bm{z})
\end{align*}

where `MLP' refers to a multi-layer perceptron with one sigmoid hidden layer.

Finally, we evaluate our representations on the relevant test portion. Whilst \citeauthor{conneau_what_2018} used grid search to determine the best hyperparameters for each probing task, we did not do so, due to both time constraints, and in an attempt to ensure classifier uniformity across languages. We describe our probing results in Section~\ref{sec:eval}.

\section{Data}
\label{sec:data}
\subsection{Probing data}
We build our probing datasets using the relevant language's Wikipedia dump as a corpus. Specifically, we use Wikipedia dumps (dated 2019-02-01), which we process using the WikiExtractor utility\footnote{\url{https://github.com/attardi/wikiextractor/}}. We use the Punkt tokeniser~\citep{kiss2006unsupervised} to segment our Wikipedia dumps into discrete sentences. For Russian, which lacked a Punkt tokenisation model, we used the UDPipe~\citep{udpipe:2017} toolkit to perform segmentation.

Having segmented our data, we used the Moses~\citep{koehn2007moses} tokeniser for the appropriate language, falling back to English tokenisation when unavailable: this was similar to XNLI's tokenisation schema, and therefore necessary for appropriate evaluation on XNLI.

Next, we obtained dependency parses for our sentences, again using the UDPipe toolkit's pretrained models, trained on Universal Dependencies treebanks~\citep{nivre_ud}. We then processed these dependency parsed corpora to extract the appropriate sentences; each task had 120k extracted sentences, divided into training/validation/test splits with 100k, 10k and 10k sentences respectively.

\begin{figure*}[t]
    \centering
    \includegraphics[width=\textwidth]{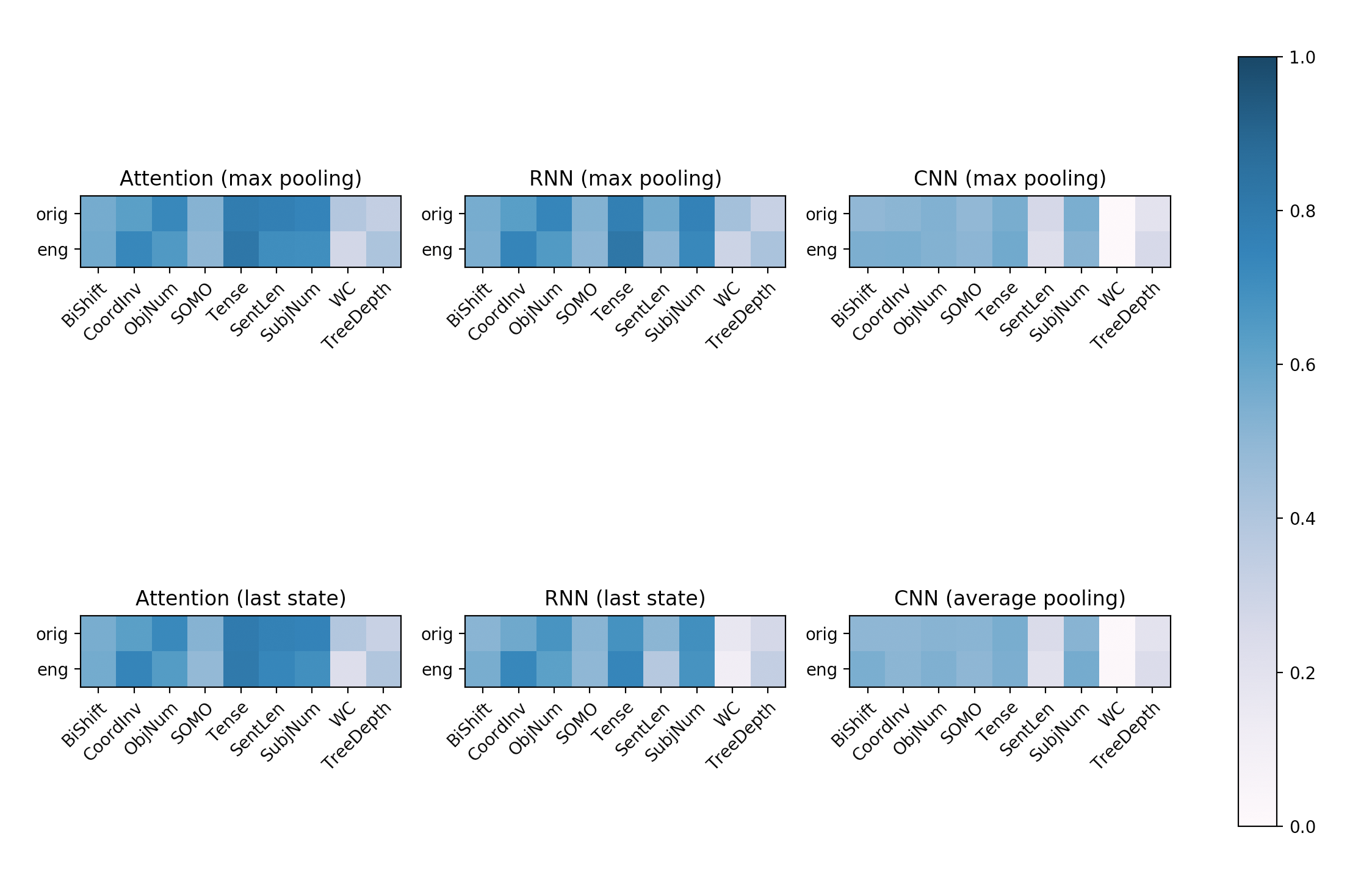}
    \caption{Probing accuracies for our six encoders on \citeauthor{conneau_what_2018}'s dataset (\texttt{orig}), compared to our Wikipedia-derived dataset (\texttt{eng})}
    \label{fig:eng}
\end{figure*}
\subsection{Mapping data}
For mapping our sentence representations, we were restricted by the availability of large parallel corpora we could use for our mapping procedure. We used two such corpora: the Europarl corpus~\citep{koehn2005europarl}, a multilingual collection of European Parliament proceedings, and the MultiUN corpus~\citep{TIEDEMANN12.463}, a collection of translated documents from the United Nations. We used Europarl for the official EU languages we analysed: German and Spanish. For Russian, we used MultiUN. We used both corpora for French, to attempt to analyse what, if any, effect the mapping corpus would have. We also truncated our MultiUN cororpora to 2 million sentences, to keep the corpus size roughly equivalent to Europarl, and also due to time and resource constraints: mapping representations on the complete 10 million sentence corpus would have required significant amounts of time.

Both our corpora were pre-segmented: we followed the same Moses-based tokenisation scheme that we did for our probing corpora, falling back to English for languages that lacked appropriate tokeniser models. 

\section{Evaluation}
\label{sec:eval}
As a preface to this section, we reiterate that the goal of this work was not to attempt to reach state-of-the-art on the tasks we describe; our goal was primarily to study the effect of transfer on sentence representations.

Our first step during evaluation, therefore, was to probe all our encoders using~\citeauthor{conneau_what_2018}'s original probing corpus, and compare these results to our English-language results on our Wikipedia-generated corpus. We present these results in the form of a heatmap in Figure~\ref{fig:eng}. 

Similarities between results on our corpora are instantly visible; these also appear to hold across encoders. Tasks with minor visible differences include \textbf{WC}, the most `difficult' classification task (1k classes), and \textbf{TreeDepth}, where we use dependency tree depth instead of constituency tree depth, as well as a different sampling mechanism.
\begin{figure}[b]
    \centering
    \includegraphics[width=\linewidth]{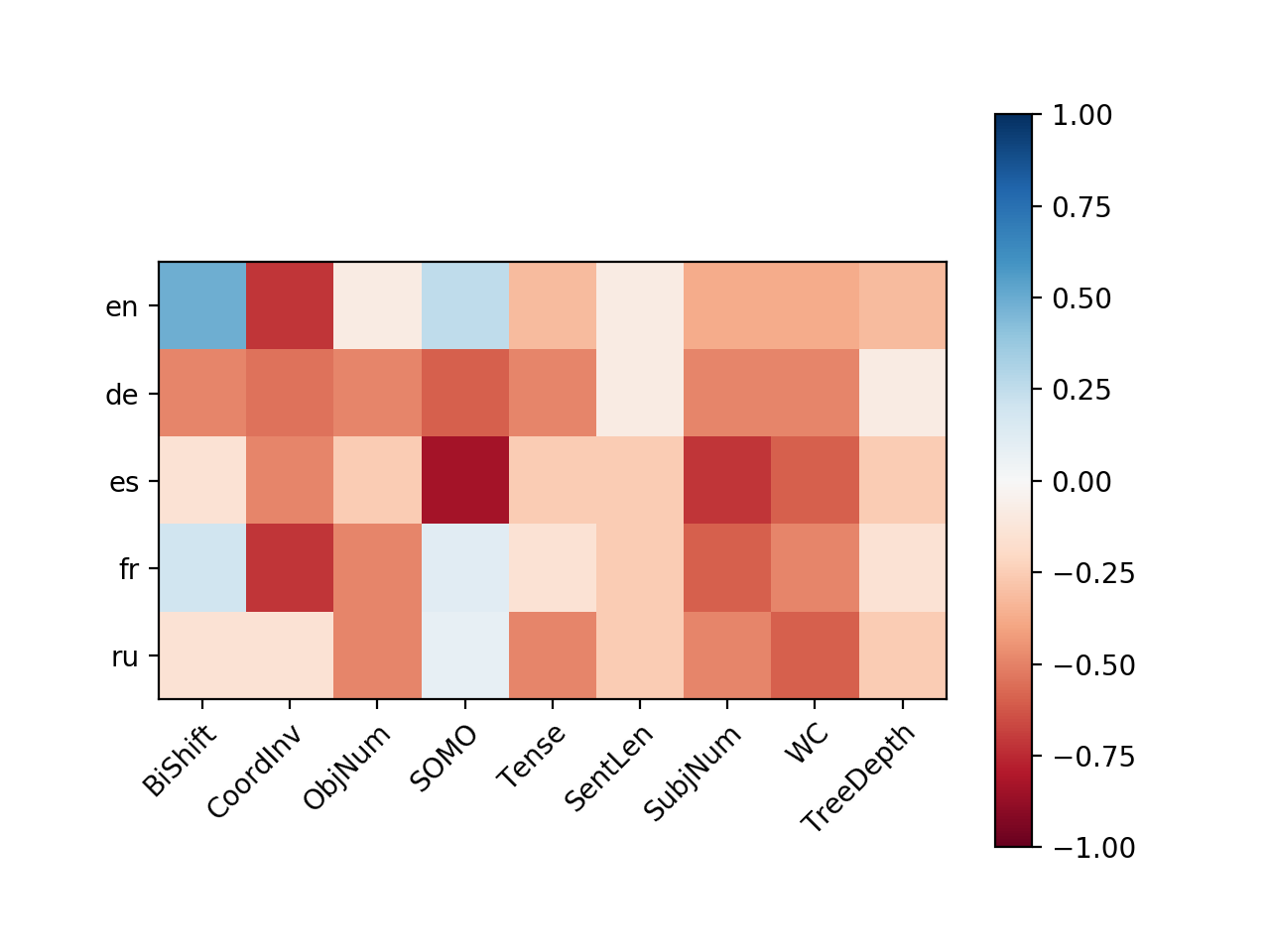}
    \caption{Spearman correlation between probing performance and XNLI; results are not statistically significant.}
    \label{fig:corel}
\end{figure}
\begin{figure*}[h!]
    \centering
    \includegraphics[width=\textwidth]{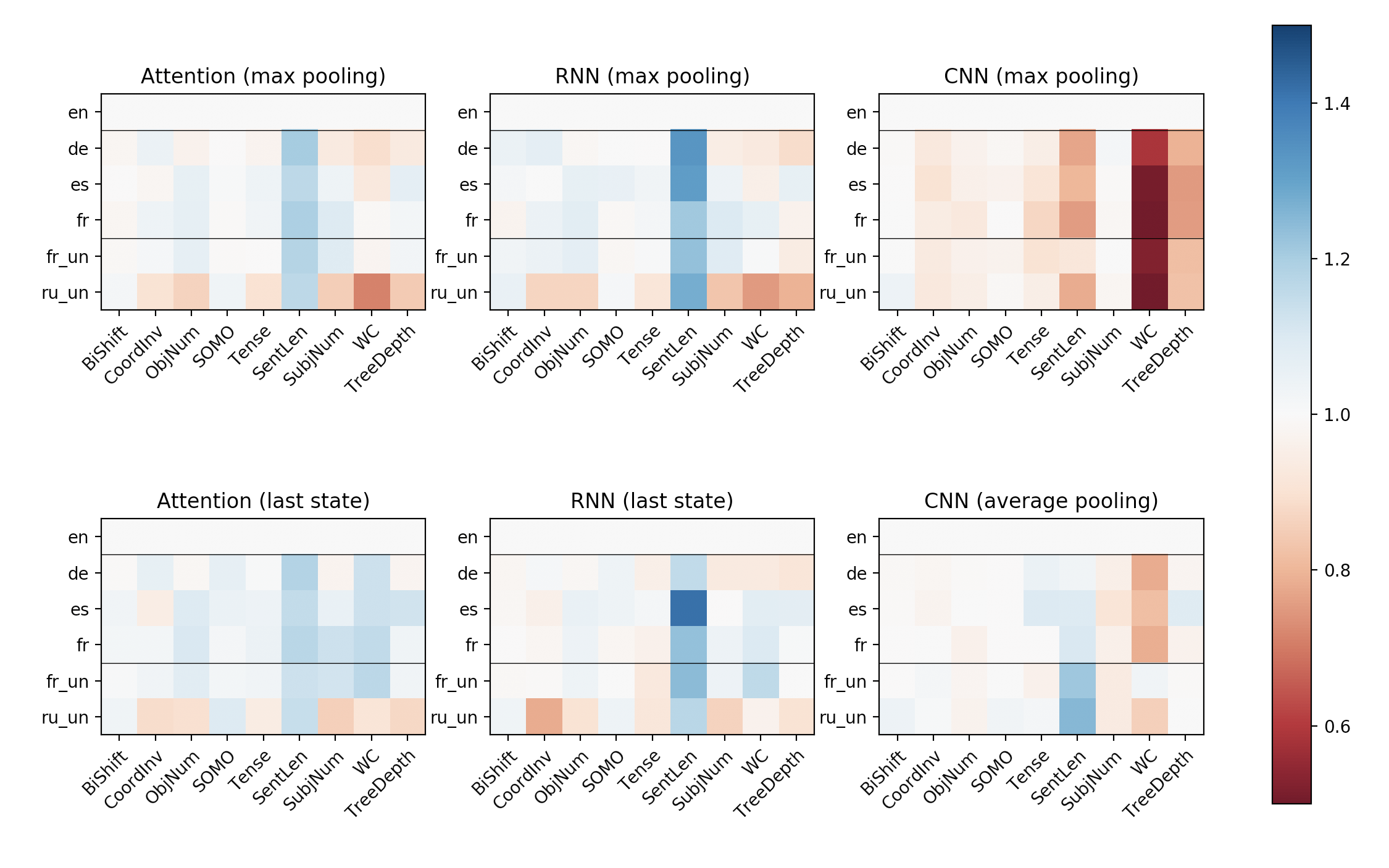}
    \caption{Probing results for each encoder relative to results on English. The second horizontal line indicates a switch in corpora. A white square indicates a value of 1, i.e. a parity in performance}
    \label{fig:cross}
\end{figure*}

Next, we present Spearman correlations between the performance of our encoders on probing tasks and on the only `true' cross-lingual downstream task we evaluated our systems on: cross-lingual natural language inference, via the XNLI~\citep{conneau_xnli:_2018} corpus. A caveat here is that we make no claims about the statistical significance of these results; given just six data points per language per task, our $p$-values tend to be well below acceptable for statistical significance. We refer the reader to \citeauthor{conneau_what_2018}'s original probing work, where despite having results for 30 encoders, correlations between many downstream and probing tasks were not statistically significant. Our correlations are presented, again in the form of a heatmap, in Figure~\ref{fig:corel}. Our absolute results on XNLI are presented in the appendix. These are not a focus for this work: we did not attempt to obtain state-of-the-art, nor, indeed, perform any sort of hyperparameter optimisation to get the `best' possible results. Given these caveats, we draw the reader's attention to the fact that the overwhelming majority of correlations are negative.

Finally, and most importantly, we measure downstream performance on probing tasks for all our cross-lingually mapped encoders. For visualisation relevant to our goals, and for brevity, we present these results, in Figure~\ref{fig:cross}, as a heatmap of probing results \emph{relative} to (our) English probing results; a full table with numeric results is presented in Appendix~\ref{sec:add}.

\section{Discussion}
\label{sec:disc}
Our cross-lingual results display some very interesting characteristics, that we enumerate and attempt to explain in this section. These results can be analysed along three dimensions: that of language, encoder mechanism, and the probing task itself.

\subsection{Language}
Whilst our results are broadly similar across languages, Russian appears to be an exception to this: our probing performance for most tasks is considerably worse when transferred to Russian than other languages. Transfer corpus does not appear to be a factor in this case: most of our encoders perform very similarly on both the Europarl and the UN variants of our transferred French representations. These are interesting preliminary results, that would require further analysis: as we mentioned in an earlier section, we were rather limited in our choice of languages, however, we foresee a possible extension to this work including more typologically diverse languages. One possible explanation for the relatively poor results on Russian is the nature of the word embeddings themselves: whilst we did not use the same methods, we did map our embeddings to the same space using the same dictionaries as~\citet{conneau_word_2017}. The results they describe for word translation retrieval are considerably poorer for English and Russian than they are for English and Spanish, French or German.

\subsection{Probing task}
An immediate surprising takeaway from our results is the (perhaps counter-intuitive) fact that transferred representations are not necessarily worse at probing tasks than trained representations are. To help with the analysis of Figure~\ref{fig:cross}, we present Table~\ref{tab:stats}, where several trends are easily visible. In particular, a task that appears to stand out is \textbf{SentLen}, with transferred encoders displaying considerably improved performance in five out of six cases.

Apart from sentence length, both number prediction tasks -- \textbf{SubjNum} and \textbf{ObjNum} -- show noticeable improvements when transferred to a non-English language. The fact that this improvement is consistent across both number tasks likely also rules out mere coincidence. We hypothesise that the explanation for these three tasks in particular showing improvements on transfer lies in the specific nature of the mapping task. While it is plausible that this is due to these specific phenomena being less critical to NLI (on which our English encoders were trained) than to the attempt made by our target encoders to \emph{emulate} these English representations, it is not immediately clear how these encoders are capable of exceeding the predictive capabilities of the encoders they are attempting to emulate.

Another interesting observation is the variance in performance for the word content (\textbf{WC}) task, which also happens to be the `hardest' task with the most output classes. We further note that, regrettably, none of our encoders were able to learn anything on \textbf{SOMO}.

\begin{table}[h]
    \centering
    \begin{tabular}{c|cc}
        Task & $\mu$ & $\sigma$  \\
        \midrule
        BiShift & 0.558 & 0.013\\
        CoordInv & 0.656 & 0.111\\
        ObjNum & 0.605 & 0.073\\
        SOMO & 0.505 & 0.011\\
        Tense & 0.708 & 0.124\\
        SentLen & 0.523 & 0.259\\
        SubjNum & 0.643 & 0.099\\
        WC & 0.152 & 0.115\\
        TreeDepth & 0.330 & 0.082\\
    \end{tabular}
    \caption{Mean and standard deviations for the absolute performance for each probing task, across languages and encoders}
    \label{tab:stats}
\end{table}
\subsection{Encoder}
All our encoders do appear to display very distinctive probing patterns, with variants of each encoder being more similar to each other than to different encoders. We enumerate some of the key observations:

\begin{enumerate}
    \item Both our CNNs appear to perform worse than attentive or recurrent mechanisms; this is, however, perfectly understandable, as our CNN-based models had an order of magnitude fewer parameters than the recurrent ones. The choice of pooling mechanism, however, appears to have a more significant effect on convolutional encoders than on others.
    \item Attentive encoders appear to be better at probing in general, whilst recurrent encoders show extremely strong performance on certain tasks, such as sentence length. 
    \item The max-pooled CNN is the only encoder that shows considerably worse performance on sentence length. This is also true for English, as is visible from Figure~\ref{fig:eng}. We hypothesise that the fixed-length filters used in convolutional encoders do not store much information about sentence length, as they only observe chunks of the sentence. A max-pooling mechanism further compounds this inability to store length by eliminating possible compositional length information that mean-pooling does ignore.
\end{enumerate}

\section{Conclusions}
\label{sec:conc}
Our analysis reveals several interesting patterns that appear to hold during cross-lingual transfer. Several of our probing tasks give us clearer insight into the sentence representations that we have generated by cross-lingual mapping, which is much needed: the principle of learning a sentence representation in parallel, combined with the fact that these representations actually appear to `work' downstream, raises a lot of questions both about what information sentence representations hold, but more interestingly, in a cross-lingual context, about what \emph{mutual} information a sentence and its translation contain.

We open-source both our training code and the probing datasets (that we dub X-P\textsc{robe})\footnote{\url{https://github.com/ltgoslo/xprobe}} that we generated in the hope that the domain of cross-lingual analysis sees further work. There are several avenues for expansion, the most obvious being a probing-oriented analysis of more complex contextualisers, such as BERT, as well as of massively multilingual or language agnostic model.

We also hypothesise that more can be said about probing with a different selection of probing tasks; indeed,~\citet{liu_linguistic_nodate} do provide a set of tasks that do not overlap with the tasks we have used. Selecting probing tasks that might tell allow us to better interpret cross-lingual modelling is another logical path one might follow. On a similar theme, an interesting research direction also involve adaptations of simple probing tasks describing linguistic phenomena to specialised architectures, for better comparison using SVCCA-style analyses~\citep{saphra_understanding_2018}.

Finally, we would also like to expand these datasets to more typologically diverse languages. A challenge in doing so is the availability of corpora that are large enough; none of our probing tasks have any sentences in common, which, given the size of each task's corpus, requires a fairly large corpus for extraction. However, this process could possibly be simplified massively by removing this mutual exclusivity requirement, which would vastly simplify the process.

\bibliography{acl2019}
\bibliographystyle{acl_natbib}

\newpage 
\appendix

\section{Appendices}
\subsection{Hyperparameters}
\label{sec:hyper}
\begin{table}[h!]
    \centering
    \begin{tabular}{c|cc}
        \textbf{Component} & \textbf{Layer} & \textbf{Value}\\
        \midrule
        Global & Embeddings & 300 (FastText)\\
               & Batch size & 10\\
               & Optimiser & Adam\\
               & Learning rate & $10^{-3}$\\
        \midrule
        RNN & biLSTM dim & 512\\
            & biLSTM layers & 2\\
            & Dropout & 10\%\\
        \midrule
        CNN & Filter sizes & (3, 4, 5)\\
            & Padding & (1, 2, 2)\\
            & Channels & 800\\
            & Projection dim & 1024\\
        \midrule
        Attention & biLSTM dim & 512\\
                  & biLSTM layers & 2\\
                  & Dropout & 10\%\\
                  & MLP dim & 150\\
                  & Activation & tanh\\
                  & Attn. heads & 60\\ 
        \midrule
        Mapper & $\lambda$ & 0.25\\
        \midrule
        Probe classifier & Hidden dim & 150\\
                         & Activation & $\sigma$\\
        
    \end{tabular}
    \caption{Hyperparameters, divided by the `component' that each layer belongs to. Note that biRNN dims are per direction.}
    \label{tab:my_label}
\end{table}

\subsection{Additional results}
\label{sec:add}

\begin{table}[h!]
    \centering
    \begin{tabular}{c||c|ccccc}
        \multirow{2}{*}{\textbf{Encoder}} & \multicolumn{6}{c}{\textbf{Language}}\\
        & English & German & Spanish & French & French (UN) & Russian\\
        \midrule
        RNN (maxpool) & \textbf{0.71} & \textbf{0.66} & 0.68 & 0.68 & 0.65 & 0.61\\
        RNN (last) & 0.66 & 0.63 & 0.65 & 0.65 & 0.63 & 0.59\\
        \midrule
        CNN (maxpool) & 0.51 & 0.39 & 0.41 & 0.36 & 0.44 & 0.43\\
        CNN (avg. pool) & 0.51 & 0.50 & 0.51 & 0.50 & 0.50 & 0.48\\
        \midrule
        Attn. (maxpool) & \textbf{0.71} & 0.64 & 0.67 & 0.67 & 0.67 & 0.60\\
        Attn. (last) & 0.70 & 0.65 & \textbf{0.69} & \textbf{0.69} & \textbf{0.66} & \textbf{0.62}\\

    \end{tabular}
    \caption{Language-specific results on relevant XNLI splits for each encoder}
    \label{tab:my_label}
\end{table}

\begin{table*}[h!]
    \centering
    \resizebox{\textwidth}{!}{\begin{tabular}{c|ccccccccc}
\textbf{English}& \textbf{BiShift} & \textbf{CoordInv} & \textbf{ObjNum} & \textbf{SOMO} & \textbf{Tense} & \textbf{SentLen} & \textbf{SubjNum} & \textbf{WC} & \textbf{TreeDepth}\\
\midrule
Attention (maxpool) & 0.57 &0.73 &0.65 &0.5 &0.82 &0.7 &0.7 &0.27 &0.41\\
Attention (last) & 0.56 &0.74 &0.64 &0.49 &0.8 &0.74 &0.7 &0.22 &0.4\\
RNN (maxpool) & 0.54 &0.74 &0.65 &0.5 &0.82 &0.51 &0.73 &0.3 &0.42\\
RNN (last) & 0.55 &0.73 &0.62 &0.5 &0.74 &0.38 &0.68 &0.11 &0.34\\
CNN (maxpool) & 0.55 &0.55 &0.53 &0.51 &0.57 &0.22 &0.52 &0.01 &0.26\\
CNN (avg. pool) & 0.55 &0.51 &0.54 &0.5 &0.54 &0.21 &0.56 &0.02 &0.24\\
\rule{0pt}{4ex}    
   \textbf{German}& \textbf{BiShift} & \textbf{CoordInv} & \textbf{ObjNum} & \textbf{SOMO} & \textbf{Tense} & \textbf{SentLen} & \textbf{SubjNum} & \textbf{WC} & \textbf{TreeDepth}\\
Attention (maxpool) & 0.56 &0.76 &0.63 &0.5 &0.8 &0.85 &0.66 &0.24 &0.39\\
Attention (last) & 0.56 &0.79 &0.63 &0.52 &0.81 &0.87 &0.68 &0.25 &0.39\\
RNN (maxpool) & 0.57 &0.8 &0.64 &0.51 &0.82 &0.68 &0.69 &0.28 &0.37\\
RNN (last) & 0.54 &0.74 &0.61 &0.52 &0.71 &0.44 &0.63 &0.11 &0.31\\
CNN (maxpool) & 0.54 &0.51 &0.51 &0.5 &0.55 &0.17 &0.53 &0.0 &0.21\\
CNN (avg. pool) & 0.54 &0.5 &0.53 &0.5 &0.57 &0.21 &0.54 &0.01 &0.23\\ 
\rule{0pt}{4ex}    
\textbf{Spanish}& \textbf{BiShift} & \textbf{CoordInv} & \textbf{ObjNum} & \textbf{SOMO} & \textbf{Tense} & \textbf{SentLen} & \textbf{SubjNum} & \textbf{WC} & \textbf{TreeDepth}\\
Attention (maxpool) & 0.57 &0.72 &0.69 &0.51 &0.85 &0.82 &0.73 &0.25 &0.44\\
Attention (last) & 0.58 &0.71 &0.7 &0.51 &0.84 &0.85 &0.74 &0.25 &0.45\\
RNN (maxpool) & 0.55 &0.75 &0.69 &0.53 &0.85 &0.67 &0.76 &0.28 &0.44\\
RNN (last) & 0.55 &0.7 &0.65 &0.52 &0.75 &0.54 &0.68 &0.12 &0.36\\
CNN (maxpool) & 0.55 &0.5 &0.51 &0.49 &0.52 &0.18 &0.51 &0.0 &0.19\\
CNN (avg. pool) & 0.55 &0.5 &0.54 &0.5 &0.6 &0.23 &0.51 &0.01 &0.26\\
\rule{0pt}{4ex}
\textbf{French}& \textbf{BiShift} & \textbf{CoordInv} & \textbf{ObjNum} & \textbf{SOMO} & \textbf{Tense} & \textbf{SentLen} & \textbf{SubjNum} & \textbf{WC} & \textbf{TreeDepth}\\
Attention (maxpool) & 0.56 &0.76 &0.7 &0.5 &0.85 &0.84 &0.76 &0.27 &0.42\\
Attention (last) & 0.58 &0.76 &0.71 &0.5 &0.84 &0.86 &0.79 &0.26 &0.41\\
RNN (maxpool) & 0.53 &0.78 &0.7 &0.5 &0.84 &0.61 &0.8 &0.31 &0.4\\
RNN (last) & 0.55 &0.72 &0.65 &0.49 &0.71 &0.47 &0.71 &0.12 &0.34\\
CNN (maxpool) & 0.55 &0.52 &0.49 &0.51 &0.5 &0.17 &0.51 &0.0 &0.2\\
CNN (avg. pool) & 0.55 &0.51 &0.52 &0.5 &0.54 &0.23 &0.54 &0.01 &0.23\\
\rule{0pt}{4ex}
\textbf{French (UN)}& \textbf{BiShift} & \textbf{CoordInv} & \textbf{ObjNum} & \textbf{SOMO} & \textbf{Tense} & \textbf{SentLen} & \textbf{SubjNum} & \textbf{WC} & \textbf{TreeDepth}\\
Attention (maxpool) & 0.57 &0.74 &0.7 &0.5 &0.82 &0.83 &0.76 &0.27 &0.42\\
Attention (last) & 0.57 &0.76 &0.69 &0.5 &0.83 &0.83 &0.78 &0.26 &0.41\\
RNN (maxpool) & 0.56 &0.78 &0.7 &0.5 &0.83 &0.62 &0.79 &0.3 &0.39\\
RNN (last) & 0.55 &0.73 &0.65 &0.5 &0.68 &0.47 &0.71 &0.13 &0.34\\
CNN (maxpool) & 0.55 &0.51 &0.51 &0.49 &0.52 &0.2 &0.52 &0.0 &0.21\\
CNN (avg. pool) & 0.55 &0.52 &0.52 &0.5 &0.52 &0.25 &0.53 &0.02 &0.24\\
\rule{0pt}{4ex}
\textbf{Russian}& \textbf{BiShift} & \textbf{CoordInv} & \textbf{ObjNum} & \textbf{SOMO} & \textbf{Tense} & \textbf{SentLen} & \textbf{SubjNum} & \textbf{WC} & \textbf{TreeDepth}\\
Attention (maxpool) & 0.58 &0.66 &0.56 &0.52 &0.74 &0.82 &0.6 &0.2 &0.35\\
Attention (last) & 0.58 &0.66 &0.57 &0.53 &0.76 &0.84 &0.6 &0.2 &0.35\\
RNN (maxpool) & 0.57 &0.65 &0.57 &0.51 &0.76 &0.65 &0.61 &0.22 &0.33\\
RNN (last) & 0.57 &0.57 &0.56 &0.52 &0.68 &0.45 &0.59 &0.11 &0.3\\
CNN (maxpool) & 0.57 &0.51 &0.5 &0.5 &0.55 &0.17 &0.51 &0.0 &0.21\\
CNN (avg. pool) & 0.57 &0.51 &0.52 &0.52 &0.56 &0.26 &0.53 &0.01 &0.24\\

    \end{tabular}}
    \caption{Complete set of absolute results per probing task, per encoder, per language. For English, these numbers are for unmapped, NLI-based encoders; for all other languages, these are post-mapping numbers}
    \label{tab:my_label}
\end{table*}

\end{document}